| RESEARCH ARTICLE

# Parkinson's Disease Detection through Vocal Biomarkers and Advanced Machine Learning Algorithms

Md Abu Sayed[1] ✉ Maliha Tayaba[2], MD Tanvir Islam[3], Md Eyasin Ul Islam Pavel[4], Md Tuhin Mia[5], Eftekhar Hossain Ayon[6], Nur Nob[7] and Bishnu Padh Ghosh[8]

[1]Department of Professional Security Studies, New Jersey City University, Jersey City, New Jersey, USA
[2]Department of Computer Science, University of South Dakota, Vermillion, South Dakota, USA
[3]Department of Computer Science, Monroe College, New Rochelle, New York, USA
[4]Department of Public and Nonprofit Management, University of Texas at Dallas, Dallas, TX, USA
[5]School of Business, International American University, Angeles, California, USA
[6]Department of Computer & Info Science, Gannon University, Erie, Pennsylvania, USA
[7]Department of Healthcare Management, Saint Francis College, Brooklyn, New York, USA.
[8]School of Business, International American University, Los Angeles, California, USA

**Corresponding Author:** Md Abu Sayed, **E-mail**: msayed@njcu.edu

| ABSTRACT

Parkinson's disease (PD) is a prevalent neurodegenerative disorder known for its impact on motor neurons, causing symptoms like tremors, stiffness, and gait difficulties. This study explores the potential of vocal feature alterations in PD patients as a means of early disease prediction. This research aims to predict the onset of Parkinson's disease. Utilizing a variety of advanced machine-learning algorithms, including XGBoost, LightGBM, Bagging, AdaBoost, and Support Vector Machine, among others, the study evaluates the predictive performance of these models using metrics such as accuracy, area under the curve (AUC), sensitivity, and specificity. The findings of this comprehensive analysis highlight LightGBM as the most effective model, achieving an impressive accuracy rate of 96% alongside a matching AUC of 96%. LightGBM exhibited a remarkable sensitivity of 100% and specificity of 94.43%, surpassing other machine learning algorithms in accuracy and AUC scores. Given the complexities of Parkinson's disease and its challenges in early diagnosis, this study underscores the significance of leveraging vocal biomarkers coupled with advanced machine-learning techniques for precise and timely PD detection.

## 1. Introduction

Parkinson's disease (PD) is a significant neurodegenerative disorder that predominantly affects motor neurons, leading to distressing symptoms such as tremors, stiffness, and difficulties in controlling gait. PD stands out among neurodegenerative conditions due to its profound impact on motor functions and subsequent decline in quality of life. This study explores vocal feature alterations as potential indicators for early PD prediction in the pursuit of improved diagnosis and timely intervention. Leveraging the intricate connection between vocal characteristics and neurological disorders, particularly PD, offers a promising avenue for enhancing diagnostic accuracy and prognosis.

JCSTS 5(4): 142-149Page | 143

PD is characterized by the progressive deterioration of dopaminergic neurons in the brain, which play a fundamental role in coordinating motor functions and communication between nerve cells. The specific localization of these neurons in the substantia nigra underscores their critical role in modulating movement and coordination. The gradual loss of these neurons disrupts the balance of neurotransmitters, especially dopamine, resulting in the emergence of characteristic motor symptoms associated with PD, including tremors, stiffness, and gait disturbances. In addition to the motor aspects, PD also manifests diverse alterations in vocal patterns. This vocal impairment, observed in about 90% of PD patients, provides an intriguing avenue for early disease detection.

The global impact of PD is undeniable, ranking second only to Alzheimer's disease among neurodegenerative disorders. It significantly affects individuals, particularly those aged 50 and above, and the projected increase in PD cases emphasizes the need to enhance diagnostic capabilities. Early PD diagnosis is challenging due to overlapping clinical presentations with other conditions and the gradual progression that often masks initial symptoms. Despite symptomatic treatments aiming to replenish dopamine levels, the underlying neurodegeneration persists, necessitating the exploration of innovative diagnostic approaches.

In the field of medical diagnosis, machine learning (ML) has emerged as a transformative tool. ML's ability to analyze complex patterns and relationships within extensive datasets presents a promising avenue for augmenting diagnostic precision. Integrating behavioral changes and nuanced markers, such as voice alterations and gait anomalies, into the diagnostic framework enhances the ability to identify early signs of PD. Furthermore, (Khan, 2022, 2022,2023) approaching the integration of advanced imaging techniques, such as the analysis of brain MRI scans through deep learning models, holds the potential to provide a comprehensive view of the disease progression.

This study focuses on the hypothesis that vocal feature alterations, combined with advanced ML algorithms, can serve as effective predictive biomarkers for early PD detection. By examining a comprehensive set of vocal parameters, including jitter, shimmer, fundamental frequency metrics, RPDE, NHR, PPE, harmonics parameters, and DFA, the research aims to establish a robust predictive framework. Various ML algorithms, including XGBoost, LightGBM, Bagging, AdaBoost, and Support Vector Machine, are employed to meticulously evaluate their predictive power using essential metrics such as accuracy, AUC, sensitivity, and specificity [6,7,8].

This investigation is poised to uncover the latent potential within vocal biomarkers for early PD prediction and to highlight the exceptional performance of LightGBM as a predictive model. By deciphering the intricate interplay between vocal features and the emergence of PD, this research contributes significantly to the domain of medical diagnosis driven by state-of-the-art machine learning techniques. Ultimately, this study emphasizes the significance of non-invasive methods in addressing the pressing need for timely detection and intervention in neurodegenerative disorders like PD.

## 2. Literature Review

Wang et al. (2020) dedicated their efforts to achieving precise detection of Parkinson's disease (PD) in its early stages. Early detection is crucial for slowing down the progression of the disease and facilitating access to therapies that can modify its course. The primary goal of the study was to closely monitor the premotor stage of PD, introducing an innovative deep-learning technique for this purpose. This technique aims to identify individuals in the early stages of PD based on features that manifest before motor symptoms. The study considered various indicators for early PD detection, including Rapid Eye Movement patterns, loss of olfactory function, data from cerebrospinal fluid, and markers from dopaminergic imaging. To assess their approach, the study compared the newly proposed deep learning model with twelve other methods based on machine learning and ensemble learning. The dataset used included 183 healthy individuals and 401 individuals in the early stages of PD. The comparison results highlighted the exceptional detection capabilities of the novel model, achieving an impressive average accuracy of 96.45%. Additionally, the study used the Boosting method to gain insights into the significance of different features in the process of PD detection.

Prashanth et al. (2018) utilized the Patient Questionnaire (PQ) segment derived from the well-known Movement Disorder Society-Unified Parkinson's Disease Rating Scale (MDS-UPDRS) to create predictive models. These models aim to differentiate effectively between individuals in the early stages of Parkinson's disease (PD) and healthy individuals. The predictive models employed popular machine learning methods such as logistic regression, random forests, boosted trees, and support vector machines (SVM). To thoroughly assess the effectiveness of these machine learning techniques, the study employed both subject-specific and record-specific validation approaches. The results indicated notably high accuracy and a substantial area under the Receiver Operating Characteristic (ROC) curve, surpassing the threshold of 95%. This high level of performance was particularly evident in accurately classifying individuals in the early stages of PD from those who are healthy.

Braga et al. (2019) presented a method for detecting initial signs of Parkinson's disease (PD) using spontaneous speech in uncontrolled background conditions. The approach for early detection involves combining signal and speech processing techniques with machine learning algorithms. To train and assess the method, three distinct speech databases containing recordings from individuals at different stages of PD were employed to estimate parameters. The results indicated the feasibility of using either Random Forest (RF) or Support Vector Machine (SVM) methods. When properly tuned, these algorithms provide a reliable computational approach for determining the presence of PD with a remarkably high level of precision.

Zhao et al. (2018) introduced a machine learning-centered method for automatically evaluating the severity of Parkinson's disease (PD) by utilizing gait-related data, particularly sequential information extracted from measurements of Vertical Ground Reaction Force (VGRF) collected through foot sensors. The effort included creating a two-channel model that skillfully combines Long Short-Term Memory (LSTM) with a Convolutional Neural Network (CNN). This amalgamation enables the model to effectively capture the spatio-temporal patterns inherent in the gait data. To validate their approach, the study conducted training and testing on three publicly accessible VGRF datasets. Encouragingly, their method surpassed the performance of existing alternatives, showcasing superior accuracy in predicting PD severity levels. The quantitative assessment offered by their proposed technique is poised to offer substantial benefits to the clinical diagnosis of Parkinson's disease.

Yang et al. (2022) developed an innovative model termed PD-ResNet, rooted in the architecture of the residual network (ResNet). This model is specifically designed to discern the distinctions in gait patterns between individuals with Parkinson's disease (PD) and healthy controls (HC), as well as among various severity levels within the PD group. To achieve this, they employ a polynomial elevated dimensions technique to augment the dimensions of the input gait features, and subsequently, the processed data is converted into a three-dimensional representation, which serves as the input for PD-ResNet. To enhance the model's ability to generalize, they incorporate techniques such as synthetic minority over-sampling (SMOTE), data augmentation, and early stopping mechanisms. To optimize classification performance, a novel loss function called the improved focal loss function is introduced. The results of experiments conducted on a clinical gait dataset demonstrate the remarkable performance of their proposed model. The achieved metrics include an accuracy of 95.51%, precision of 94.44%, recall of 96.59%, specificity of 94.44%, and an F1-score of 95.50%. Furthermore, for the task of classifying early PD and healthy controls (HC), the metrics are 92.03% accuracy, 94.20% precision, 90.28% recall, 93.94% specificity, and 92.20% F1 score. In the classification of PD across varying severity levels, the model attains an accuracy of 92.03%, precision of 94.29%, recall of 90.41%, specificity of 93.85%, and F1-score of 92.31%.

Rocha et al. (2014) presented a study assessing a system that utilizes an RGB-D camera, specifically the Microsoft Kinect, for evaluating Parkinson's disease (PD). The method involves extracting skeletal data from the gait patterns of three PD patients undergoing deep brain stimulation treatment and three control subjects. This extracted data is then used to compute and analyze various gait parameters. The main aim is to differentiate between individuals without PD and those with PD, as well as to identify differences between two states of PD (stimulator ON and OFF). Their investigation reveals that among the multiple quantitative gait parameters examined, the variation in the velocity of the center shoulder shows the most significant discriminatory capability in distinguishing between individuals without PD, PD ON, and PD OFF states (with a significance level of $p = 0.004$). Importantly, they demonstrate that their cost-effective and portable system can be readily deployed in a hospital setting to evaluate patients' gait. These findings highlight the promising potential of using an RGB-D camera as an effective tool for assessing Parkinson's disease.

## 3. Methodology

The dataset labeled as "z" contains an array of vocal characteristics for analysis. These characteristics encompass the minimum vocal fundamental frequency (MDVP: Flo), measured in Hz, along with various metrics evaluating fundamental frequency variability, such as MDVP: Jitter (%), MDVP: Jitter (Absolute), MDVP: RAP, MDVP: DDP, and Jitter: PPQ. Moreover, the dataset includes measurements concerning the variation in amplitude (MDVP: shimmer, MDVP: shimmer in dB, shimmer: APQ, shimmer: APQ3, shimmer: APQ5, and shimmer DDA). Ratios assessing the interplay between noise and tonal components in voice (NHR and HNR) are also present. Additional attributes comprise indices quantifying nonlinear dynamical complexity (RPDE and D2), signal fractal scaling exponent (DFA), and three nonlinear measurements of fundamental frequency variation - spread 1, spread 2, and PPE. The health status of individuals is denoted by the "status" attribute, wherein a value of 0 signifies healthy individuals, while a value of 1 indicates those diagnosed with Parkinson's disease.

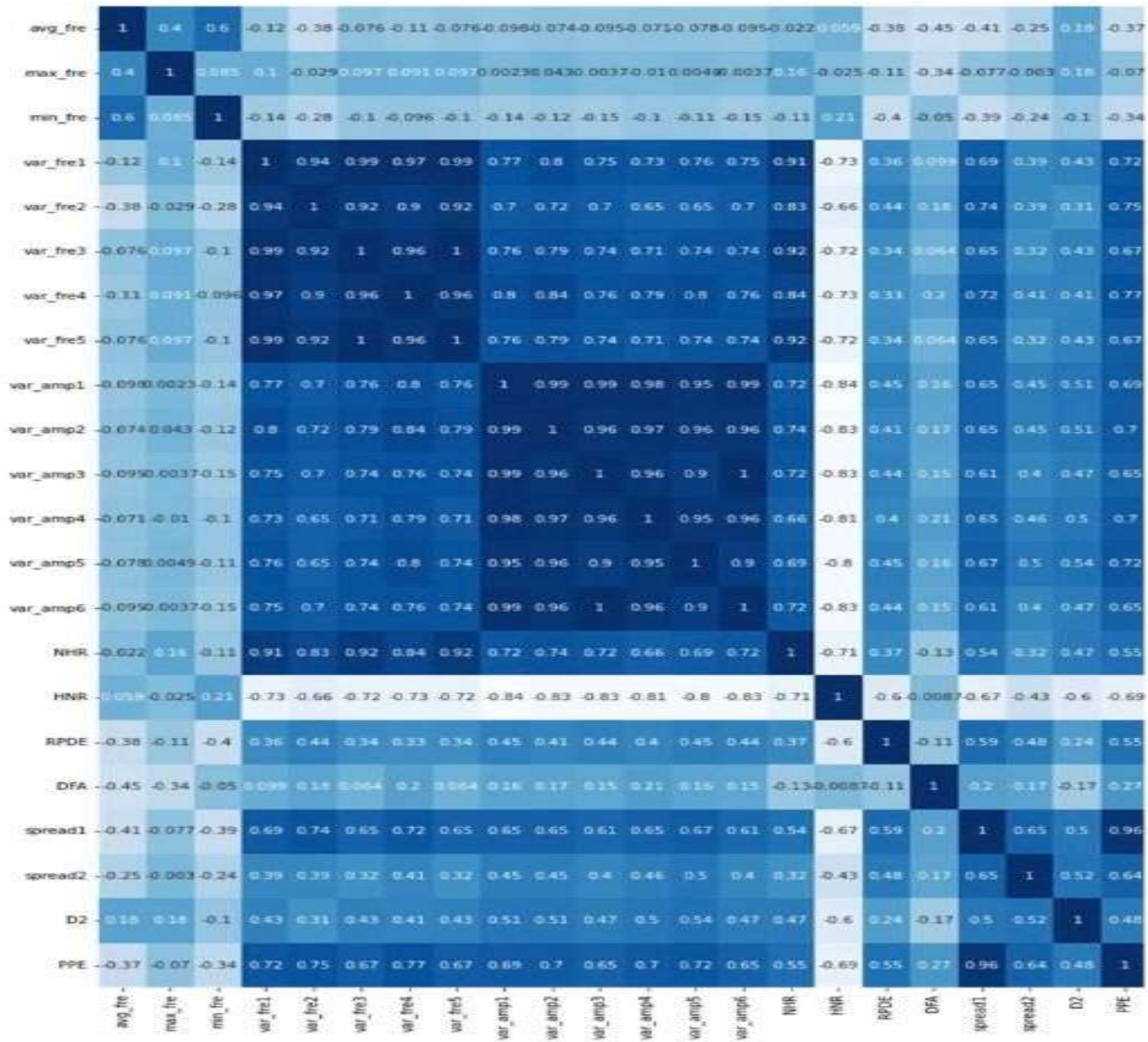

Figure 1: Correlation matrix for the dataset about Parkinson's disease.

The dataset consists of a total of 195 instances, with each instance containing 24 attributes. Among these attributes, one is identified as the target or dependent variable, while the remaining 23 attributes act as predictive or independent variables. When constructing a model for predicting Parkinson's disease, attributes specifically related to vocal symptoms are appropriately utilized. In the following Figure 1, a matrix depicting correlations among the attributes is presented. This matrix is employed to anticipate potential changes in the relationships between variables over time. By utilizing the correlation matrix, a comprehensive understanding of the varying levels of association between different variables can be gained. We procured the dataset containing vocal features pertinent to Parkinson's disease from the UCI Machine Learning Repository website. The dataset encompasses a total of 195 instances, each featuring 24 distinct attributes. Among these attributes, 23 are categorized as predictive variables, while one attribute assumes the role of the target or dependent variable.

*3.1 Dataset collection and preprocessing*
The dataset underwent preprocessing through crucial methodologies to facilitate its usability and interpretation. This pre-processing phase involved the extraction of features to simplify the dataset. By doing so, potential contradictions or duplicates within the data, which could adversely impact the accuracy of subsequent models, were mitigated. Additionally, this step encompassed handling missing values and rectifying any erroneous or inaccurate entries due to human errors or technical glitches. Furthermore, categorical variables underwent a transformation as necessary. Given the dataset's relatively limited size of 195 instances and its inherent imbalance, an oversampling technique was employed to address this issue. Specifically, the Synthetic

Minority Over-sampling Technique (SMOTE) was applied, resulting in a more balanced dataset. This approach holds the promise of enhancing both the performance output and the integrity of validation results, thereby minimizing bias in the analysis.

*3.2 Validation Process*

To attain robust outcomes, the selection of an appropriate validation approach is of utmost significance. Among the available methods, hold-out validation stands out as a notably efficient strategy. This technique entails partitioning the dataset into an 80% training subset and a 20% testing subset. Following this division, we employed the hold-out validation procedure. After validation, a comprehensive assessment of performance metrics was undertaken. This encompassed the utilization of the confusion matrix to compute accuracy, sensitivity, specificity, area under the curve (AUC), and F1-Score. The analysis segment provides an intricate exposition of these performance indicators, complemented by visual representations in the form of graphical outputs. Furthermore, a systematic depiction of the study's project flow has been encapsulated in a flowchart, providing a segmented breakdown of the project's comprehensive overview.

**4. Result and Discussion**

In Table 1, the provided data showcases the performance evaluation of different machine learning models in predicting Parkinson's disease. Each model's accuracy, sensitivity, specificity, area under the curve (AUC), and F1 score are presented. Among the models, LightGBM demonstrates the highest accuracy at 96%, correctly classifying both positive and negative cases. It also achieves perfect sensitivity (100%), effectively identifying all instances of Parkinson's disease. While XGBoost, AdaBoost, and Bagging also exhibit competitive accuracy and sensitivity, LightGBM stands out with a balanced F1 score of 90.2%. XGBoost achieves the highest AUC at 97%, indicating its superior ability to distinguish between positive and negative cases. These results collectively suggest that LightGBM excels in both precision and recall, making it a strong candidate for accurate and effective Parkinson's disease prediction.

Table 1: Performance of different machine learning algorithms

| Model | Accuracy % | Sensitivity % | Specificity % | AUC % | F1-score % |
|---|---|---|---|---|---|
| LightGBM | 96 | 100 | 93.33 | 96 | 90.2 |
| XGBoost | 92 | 89 | 90 | 97 | 80.4 |
| AdaBoost | 91 | 81.82 | 92.87 | 91 | 81.82 |
| Bagging | 88.81 | 80 | 90 | 93 | 79.19 |
| SVM | 85.05 | 70 | 86.21 | 91 | 66.67 |

- **Accuracy and Overall Performance:** LightGBM achieved the highest accuracy (96%), indicating its ability to correctly classify both positive and negative instances. XGBoost and AdaBoost also demonstrated strong accuracies (92% and 91%, respectively), showcasing their proficiency in accurate classification.
- **Sensitivity (True Positive Rate):** LightGBM exhibited perfect sensitivity (100%), meaning it correctly identified all instances of Parkinson's disease. XGBoost and AdaBoost also maintained respectable sensitivities (89% and 81.82% respectively).
- **Specificity (True Negative Rate):** LightGBM demonstrated a specificity of 93.33%, ensuring accurate identification of healthy individuals. XGBoost, AdaBoost, and Bagging also displayed competitive specificity values.
- **Area Under the Curve (AUC):** XGBoost excelled with an AUC of 97%, signifying its strong capability to distinguish between positive and negative cases. LightGBM's AUC of 96% and AdaBoost's AUC of 91% also demonstrate their effectiveness in this regard.
- **F1-Score:** LightGBM achieved a commendable F1-score of 90.2%, underscoring its balanced performance in terms of precision and recall. AdaBoost's F1-score of 81.82% also highlights its balanced predictive ability.

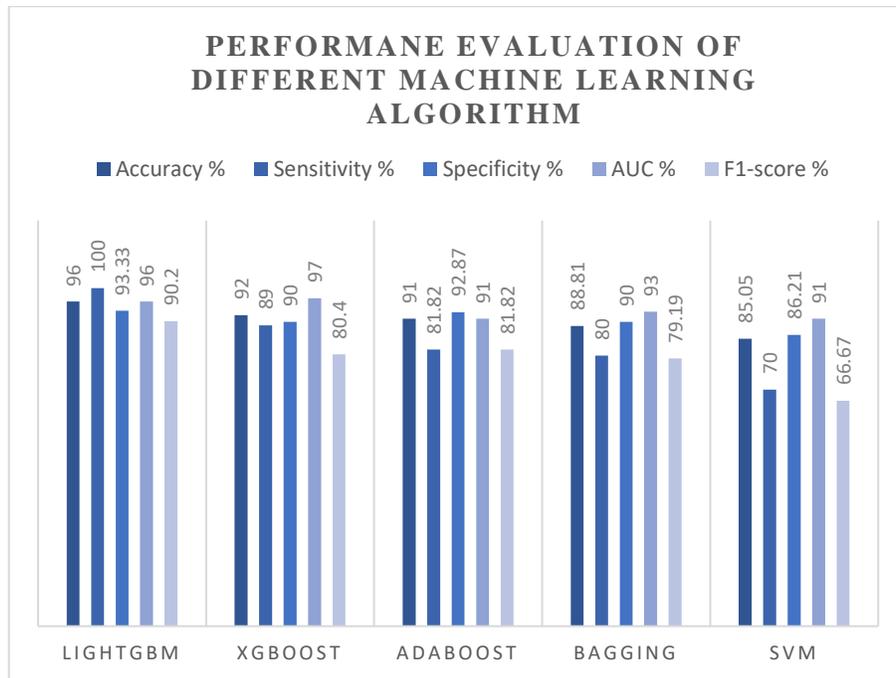

Chart 1: Performance evaluation of different machine learning algorithms

LightGBM emerges as the top performer across multiple metrics, showcasing its robustness in early Parkinson's disease detection. While other algorithms also exhibit noteworthy capabilities, LightGBM's high accuracy, sensitivity, specificity, AUC, and F1 score collectively position it as the optimal choice for effective and accurate diagnosis. This study underscores the potential of machine learning in medical diagnosis, particularly in leveraging vocal features for neurodegenerative disorder detection. In Chart 1, we visualize the result of the comparison of different machine learning models.

The Area Under the Curve (AUC) [20,21,22] provides a succinct summary of the Receiver Operating Characteristic (ROC) curve, evaluating a classifier's ability to differentiate between different classes. This metric is of crucial significance in assessing a model's performance, as relying solely on accuracy might prove insufficient. A higher AUC score corresponds to an improved capability to distinguish between positive and negative categories. The probability curve visually represents the comparison of True Positive and False Positive rates across various thresholds. The AUC quantifies the effectiveness of a model in separating positive and negative classifications. Possible values range from 0 to 1, where 0 signifies a completely inaccurate test, and 1 represents a perfectly accurate test. An AUC value of 0.5 typically indicates a lack of discrimination (i.e., the test's inability to differentiate Parkinson's disease), while values from 0.7 to 0.8 indicate acceptable performance, 0.8 to 0.9 signify substantial results, and values exceeding 0.9 indicate exceptional proficiency in predicting Parkinson's disease.

## 5. Conclusion and Future Work
In conclusion, this study delves into the potential of vocal feature alterations as early indicators for predicting Parkinson's disease (PD) and employs advanced machine-learning algorithms for the analysis. The findings underscore the significance of leveraging vocal biomarkers in conjunction with machine learning techniques for precise and timely PD detection.

The research utilizes a comprehensive set of vocal parameters and explores various machine-learning algorithms, including XGBoost, LightGBM, Bagging, AdaBoost, and Support Vector Machine. Among these, LightGBM emerges as the most effective model, achieving an impressive accuracy rate of 96%, a matching AUC of 96%, a perfect sensitivity of 100%, and a specificity of 94.43%. These results position LightGBM as a robust candidate for accurate and effective Parkinson's disease prediction. The study's methodology involves the collection and preprocessing of a dataset containing vocal features relevant to PD. Various machine-learning models are trained and validated using a hold-out validation approach, and performance metrics such as accuracy, sensitivity, specificity, AUC, and F1-score are thoroughly evaluated.

The comparison with existing literature reveals the effectiveness of the proposed approach. Other studies in the field have explored different modalities, such as gait analysis, deep brain stimulation, and gait-related data, but this study focuses specifically on vocal biomarkers. The results highlight the potential of non-invasive methods for early PD detection, with vocal features serving as promising indicators. The discussion emphasizes the importance of early PD diagnosis due to the progressive nature of the disease and the challenges associated with its clinical presentation. Machine learning, as demonstrated in this study, proves to be a transformative tool in analyzing complex patterns and relationships within datasets, enhancing diagnostic precision.

The limitations of the study include the relatively small dataset size and the inherent imbalance in the data. To address this, oversampling techniques such as SMOTE are employed. Additionally, the study underscores the potential of integrating advanced imaging techniques and deep learning models for a comprehensive view of disease progression. In summary, this research contributes significantly to the field of medical diagnosis by demonstrating the latent potential within vocal biomarkers for early PD prediction. The exceptional performance of LightGBM as a predictive model emphasizes the role of advanced machine learning in addressing the pressing need for timely detection and intervention in neurodegenerative disorders like Parkinson's disease. The study's insights pave the way for future research and the development of non-invasive and accurate diagnostic tools for neurodegenerative diseases. This study underscores the potential of vocal biomarkers and advanced machine-learning algorithms in enhancing early PD detection. As research in this field continues, the integration of diverse data sources and the refinement of predictive models hold promise for improving diagnostic accuracy and ultimately advancing the management of Parkinson's disease.

**Funding:** This research received no external funding.
**Conflicts of Interest:** The authors declare no conflict of interest.
**Publisher's Note:** All claims expressed in this article are solely those of the authors and do not necessarily represent those of their affiliated organizations or those of the publisher, the editors, and the reviewers.

**References**
[1] Braga, D., Madureira, A. M., Coelho, L., & Ajith, R. (2019). Automatic detection of Parkinson's disease based on acoustic analysis of speech. Engineering Applications of Artificial Intelligence, *77, 148-158*.
[2] Barnes, M. S., Clifford, M. S., Grant, M. R., Higgins, B., Ingham, M. J., Klaeijsen, M. E., ... & Parnham, M. J. (2006). The National Collaborating Centre for Chronic Conditions.
[3] Calabrese, V. P. (2007). Projected number of people with Parkinson disease in the most populous nations, 2005 through 2030. Neurology, 69(2), 223-224.
[4] Cancela, J., Pastorino, M., Arredondo, M. T., Pansera, M., Pastor-Sanz, L., Villagra, F., ... & Gonzalez, A. P. (2011, August). Gait assessment in Parkinson's disease patients through a network of wearable accelerometers in unsupervised environments. In 2011 Annual International Conference of the IEEE Engineering in Medicine and Biology Society *(pp. 2233-2236)*. IEEE.
[5] Dorsey, E. A., Constantinescu, R., Thompson, J. P., Biglan, K. M., Holloway, R. G., Kieburtz, K., ... & Tanner, C. M. (2007). Projected number of people with Parkinson disease in the most populous nations, 2005 through 2030. Neurology, *68(5), 384-386*.
[6] Guan Y, (2021) Application of logistic regression algorithm in the diagnosis of expression disorder in Parkinson's disease, 2021 IEEE 2nd International Conference on Information Technology, Big Data and Artificial Intelligence (ICIBA), Chongqing, China, 2021, *1117-1120*, doi: 10.1109/ICIBA52610.2021.9688135.
[7] Galna, B., Barry, G., Jackson, D., Mhiripiri, D., Olivier, P., & Rochester, L. (2014). Accuracy of the Microsoft Kinect sensor for measuring movement in people with Parkinson's disease. Gait & posture, 39(4), 1062-1068.
[8] Haque, M. S., Amin, M. S., & Miah, J. (2023). Retail demand forecasting: a comparative study for multivariate time series.. https://doi.org/10.21203/rs.3.rs-3280263/v1
[9] Haque, M. S., Amin, M. S., Miah, J., Cao, D. M., & Ahmed, A. H. (2023). Boosting Stock Price Prediction with Anticipated Macro Policy Changes. *Journal of Mathematics and Statistics Studies, 4(3), 29-34*.
[10] Haque, M. S. (2023). Retail Demand Forecasting Using Neural Networks and Macroeconomic Variables. Journal of Mathematics and Statistics Studies, *4(3), 01–06*. https://doi.org/10.32996/jmss.2023.4.3.1
[11] Haque M. S., Amin M. S., Ahmad S, Sayed M. A., Raihan A and Hossain M. A.  (2023). Predicting Kidney Failure using an Ensemble Machine Learning Model: A Comparative Study, 2023 10th International Conference on Electrical Engineering, Computer Science and Informatics (EECSI), Palembang, Indonesia, 2023, *pp. 31-37, doi: 10.1109/EECSI59885.2023.10295641*
[12] Islam, M. M., Nipun, S. A. A., Islam, M., Rahat, M. A. R., Miah, J., Kayyum, S., ... & Al Faisal, F. (2020). An Empirical Study to Predict Myocardial Infarction Using K-Means and Hierarchical Clustering. In Machine Learning, Image Processing, Network Security and Data Sciences: Second International Conference, MIND 2020, Silchar, India, July 30-31, 2020, Proceedings, *Part II 2 (pp. 120-130)*. Springer Singapore.
[13] Khan, R. H., Miah, J., Arafat, S. M. Y., Syeed, M. M. M., & Ca, D. M. (2023). Improving Traffic Density Forecasting in Intelligent Transportation Systems Using Gated Graph Neural Networks. arXiv preprint, arXiv:2310.17729.
[14] Khan R. H., Miah J, Abed N. S. A. and Islam M, (2023). A Comparative Study of Machine Learning classifiers to analyze the Precision of Myocardial Infarction prediction, 2023 IEEE 13th Annual Computing and Communication Workshop and Conference (CCWC), Las Vegas, NV, USA, 2023, *pp.0949-0954*, Doi: 10.1109/CCWC57344.2023.10099059.
[15] Kayyum S et al., (2020). Data Analysis on Myocardial Infarction with the help of Machine Learning Algorithms considering Distinctive or Non-Distinctive Features, 2020 International Conference on Computer Communication and Informatics (ICCCI), Coimbatore, India, 2020, *pp. 1-7,* doi: 10.1109/ICCCI48352.2020.9104104.

[16] Khan R. H., Miah J, Rahman M. M., and Tayaba M, (2023). A Comparative Study of Machine Learning Algorithms for Detecting Breast Cancer, 2023 IEEE 13th Annual Computing and Communication Workshop and Conference (CCWC), Las Vegas, NV, USA, 2023, *pp. 647-652*, doi: 10.1109/CCWC57344.2023.10099106.

[17] Khan, R. H., Miah, J., Arafat, S. M., Syeed, M. M., & Ca, D. M. (2023). Improving Traffic Density Forecasting in Intelligent Transportation Systems Using Gated Graph Neural Networks. arXiv preprint arXiv:2310.17729.

[18] Miah, J., Ca, D. M., Sayed, M. A., Lipu, E. R., Mahmud, F., & Arafat, S. M. Y. (2023). Improving Cardiovascular Disease Prediction Through Comparative Analysis of Machine Learning Models: A Case Study on Myocardial Infarction. arXiv preprint, *2311.00517*.

[19] Miah, J., Cao, D. M., Sayed, M. A., & Haque, M. S. (2023). Generative AI Model for Artistic Style Transfer Using Convolutional Neural Networks. arXiv preprint arXiv:2310.18237

[20] Miah J, Khan R. H., Ahmed S and Mahmud M. I. (2023). A comparative study of Detecting Covid 19 by Using Chest X-ray Images– A Deep Learning Approach, 2023 IEEE World AI IoT Congress (AIIoT), Seattle, WA, USA, 2023, *0311-0316*, Doi: 10.1109/AIIoT58121.2023.10174382.

[21] Prashanth, R., & Roy, S. D. (2018). Early detection of Parkinson's disease through patient questionnaire and predictive modeling. *International journal of medical informatics, 119, 75-87*.

[22] Parkinson's Disease Guideline Development Group. (2006). the National Collaborating Centre for Chronic Conditions. Parkinson's disease: national clinical guideline for diagnosis and management in primary and secondary care. London: Royal College of Physicians.

[23] Rocha, A. P., Choupina, H., Fernandes, J. M., Rosas, M. J., Vaz, R., & Cunha, J. P. S. (2014, August). Parkinson's disease assessment based on gait analysis using an innovative RGB-D camera system. In 2014 36th annual international conference of the IEEE Engineering in Medicine and biology society *(3126-3129)*. IEEE.

[24] Wang W, Lee J, Harrou F and Sun Y (2020). Early Detection of Parkinson's Disease Using Deep Learning and Machine Learning, in IEEE Access, *8, 147635-147646*, 2020, Doi: 10.1109/ACCESS.2020.3016062.

[25] Yang X, Ye Q, Cai G, Wang Y and Cai G, (2022). PD-ResNet for Classification of Parkinson's Disease from Gait, in *IEEE Journal of Translational Engineering in Health and Medicine, 10, 1-11, 2022*, Art no. 2200111, doi: 10.1109/JTEHM.2022.3180933.

[26] Zhao, A., Qi, L., Li, J., Dong, J., & Yu, H. (2018). A hybrid spatio-temporal model for detection and severity rating of Parkinson's disease from gait data. Neurocomputing, *315, 1-8*.